\documentclass[preprint,11pt]{article}

\usepackage[ruled,linesnumbered,vlined]{algorithm2e}
\usepackage{algorithmic}
\usepackage{lipsum}
\usepackage{adjustbox}

\usepackage{graphics}
\usepackage{amssymb}
\usepackage{rotating}
\usepackage{graphicx}
\usepackage{tabularx}
\usepackage[english]{babel}
\usepackage{mathrsfs} 
\usepackage{amsmath}
\usepackage{fullpage}
\usepackage{tabularx}
\usepackage{multirow}
\usepackage{graphicx}
\usepackage{wrapfig}
\usepackage{float}
\usepackage{verbatim}
\usepackage{longtable}
\usepackage{booktabs}
\usepackage{hyperref}
\title{A new constraint programming model and a linear programming-based adaptive large neighborhood search for the vehicle routing problem with synchronization constraints}

\author{Minh Ho\`ang H\`a, Tat Dat Nguyen, Thinh Nguyen Duy, Hoang Giang Pham, Thuy Do, Louis-Martin Rousseau}

\begin{document}

\begin{center}

\vspace*{-0.5cm}

\begin{huge}
A new constraint programming model and a linear programming-based adaptive large neighborhood search for the vehicle routing problem with synchronization constraints
\end{huge}

\vspace*{0.45cm}

\textbf{Minh Ho\`ang H\`a*, Tat Dat Nguyen} \\
ORLab, VNU University of Engineering and Technology, Hanoi, Vietnam \\ 
\vspace*{0.15cm}
\textbf{Thinh Nguyen Duy, Hoang Giang Pham, Thuy Do} \\
Department of Computer Science and Operations Research, Universit\'e de Montr\'eal and CIRRELT, Montr\'eal, Qu\'ebec, Canada \\
\vspace*{0.15cm}
\textbf{Louis-Martin Rousseau} \\
\'Ecole Polytechnique de Montr\'eal and CIRRELT, Montr\'eal, Canada \\
\vspace*{0.15cm}

\end{center}
\noindent
\textbf{Abstract.}
We consider a vehicle routing problem which seeks to minimize cost subject to time window and synchronization constraints. In this problem, the fleet of vehicles is categorized into regular and special vehicles. Some customers require both vehicles' services, whose starting service times at the customer are synchronized. Despite its important real-world application, this problem has rarely been studied in the literature. To solve the problem, we propose a Constraint Programming (CP) model and an Adaptive Large Neighborhood Search (ALNS) in which the design of insertion operators is based on solving linear programming (LP) models to check the insertion feasibility. A number of acceleration techniques is also proposed to significantly reduce the computational time. The computational experiments show that our new CP model finds better solutions than an existing CP-based ANLS, when used on small instances with 25 customers and with a much shorter running time. Our LP-based ALNS dominates the cp-ALNS, in terms of solution quality, when it provides solutions with better objective values, on average, for all instance classes. This demonstrates the advantage of using linear programming instead of constraint programming when dealing with a variant of vehicle routing problems with relatively tight constraints, which is often considered to be more favorable for CP-based methods. 
\vspace*{0.2cm}


\section{Introduction}
Most of the requirements related to our daily routines are made by a service provider coming to our premises. These types of services can be home care delivery, maintenance operations, public utilities, etc. In such services, efficient delivery and timely service play important roles. This is why the class of the Vehicle Routing Problem, coupled with the Scheduling Problem, comprises a large class of problems with many variations and applications. The main focus of this research is to study the Vehicle Routing Problem with Synchronization Constraints (VRPSC), where both time window and synchronization constraints are present. In the latter constraint, some customers may require the service of two vehicles whose starting service times at the customer must be synchronized.  

In a recent industrial project with an Internet Service Provider (ISP) in Vietnam, the authors witnessed several contexts where synchronization constraints arose. The ISP company has to perform installation services for new subscribers and maintenance services for subscribed clients. Both services are performed by technicians who mainly use motorbikes for travelling. In many cases, a customer asks for services by two technicians belonging to two different teams: one being the "physical" team, which takes charge of the hardware (cable wire, modem, etc.); while the other team manages the signal. To further illustrate the problem, when a customer requires a service from the physical team, two technicians must be mobilized as one helps the other with equipment set-up, such as installing a ladder and other protective gear. In addition, an intern technician, who is in a probationary period, needs to be coupled with an experienced technician at customer locations. When servicing a customer, the company requires that the starting server time of both technicians be as close as possible in order to reduce their waiting time and to limit the disruption to the customer. In the case of our ISP partner, a delay of no more than 15 minutes is permitted. The problem was initially introduced in \cite{Hojabri} and can be used to model other real world applications such as home care delivery, aircraft fleet assignment, ground handling, and forest operations (see \cite{rousseau} for more information).

Hojabri et al. \cite{Hojabri} proposed a constraint programming-based Adaptive Large Neighborhood Search (cp-ALNS) with insertion operators exploiting constraint propagation capabilities to guarantee the feasibility of a new generated solution. Different from the popular ALNS proposed in \cite{Ropke} to solve VRPs, the cp-ALNS does not try to add all unserved requests one by one but, rather, adds all of them at once to create a new complete solution. Several removal operators were specifically designed for the problem. Numerical results are reported on instances derived from Vehicle Routing Problem with Time Window (VRPTW) benchmark instances, with up to 200 customers and 100 synchronizations.

A dynamic version of the problem was introduced in \cite{rousseau} where customer requests were not foreseen, but arrived one-by-one in real time. The problem was modeled as a CP program from which a metaheuristic was designed. We note that the methods proposed in \cite{rousseau, Hojabri} were based on the same CP model which uses variables representing the successor of a node, $AllDifferent$ and subtour elimination constraints introduced in \cite{Pesant}. However, no result of the pure CP model has been reported.  

Apart from the paper of \cite{Hojabri}, a problem quite similar to the VRPSC is studied in \cite{bredstrom} in the context of home care crew scheduling. The problem is first formulated as a mixed integer programming (MIP) model that constitutes pairwise synchronization and pairwise temporal precedence between customer visits. It is then solved through an optimization-based heuristic. Rasmussen et al. \cite{rasmussen} solves a similar problem with an exact branch-and-price algorithm. Due to the application context, there is no capacity constraint and specific issues about home care crews are taken into account, in addition to the synchronization requirements, like care giver preferences, customer priority and the ability of a particular care giver to serve a given customer. Also, not all customers must be serviced because visits can be rescheduled or canceled.

Afifi et al. \cite{afifi} propose a simulated annealing algorithm with dedicated local searches (SA-ILS) for the VRPTW with synchronized visits. Their problem differs from the considered problem in this work in two ways; first, the synchronized nodes must be serviced at the same time, i.e., the delay is zero and the fleet of vehicles is not categorized. Second, three objective functions have been considered: minimizing the total travel time; minimizing the sum of negative preferences; and minimizing the maximum difference in service times of the vehicles. Recently, Parragh et al. \cite{parragh} evaluate several different ways to deal with pairwise synchronization constraints in the context of two problems: the VRPTW with pairwise synchronization, and the service technician routing and scheduling problem. They propose three ways to address the synchronization requirement: individual synchronized timing optimization; global synchronized timing optimization; and adaptive time window. The idea of the first two approaches is to keep the ALNS untouched, while the last one makes some modifications to the insertion scheme in order to identify good starting service times for synchronized visits.

Pillac et al. \cite{pillac} introduce a Dynamic Technician Routing and Scheduling Problem (D-TRSP) which deals with a limited crew of technicians serving a set of requests. In the D-TRSP, each technician has a set of skills, tools and spare parts required by each request. In addition to designing a route at the beginning of each day, two types of decisions must be managed in real time. First, whenever a new request appears, we must decide whether it is accepted or not. And second, whenever a technician finishes serving a request, we need to find the next request to serve. For a survey on further synchronization issues in the context of vehicle routing problems, we refer the readers to \cite{drexl}.

Our main contributions are as follows: we propose a new CP model and a metaheuristic based on ALNS to address the VRP with time windows and synchronization constraints. Different from the CP model in the literature, our CP uses \textit{sequence} and \textit{interval} variables of IBM ILOG CP Optimizier \cite{CPO} to formulate the problem. The most notable feature of our metaheuristic is we use linear programming and a number of acceleration techniques to quickly check the feasibility of insertion operations integrated in the popular ALNS \cite{Ropke}. This is the first time LP models are used instead of CP to design the ALNS algorithm for the VRP with synchronization constraints. The computational experiments carried on the benchmark data show the performance of our methods. More precisely, our CP model can provide better solutions on small-size instances than the cp-ALNS of \cite{Hojabri} in a much shorter running time. Our lp-ALNS dominates the cp-ALNS in terms of solution quality, as it improves 620 over 681 best known solutions. The remainder of the paper is organized as follows: Section \ref{def} introduces the problem definition and our new CP model; the detailed description of the lp-ALNS is provided in Section \ref{method}; experimental results are reported in Section \ref{result}; and finally, we conclude our work in Section \ref{conclude}.

\section{Problem definition and a new constraint programming model}
\label{def}

The problem may be formally defined as follows: given $G = (V, E)$, an undirected graph, where $V = \{v_0\} \cup V_s \cup V_r$ is the set of nodes representing customer locations and $E$ is the set of edges. $v_0$ is the depot where a set of vehicles $M$ is located. Vehicles in set $M$ are again divided into two sub-sets: regular vehicles (set $M_1$) and special vehicles (set $M_2$). All regular vehicles have a capacity of $Q$. $V_r$ is the set of regular customers which are visited by regular vehicles only while $V_s$ is the set of special customers, each requires the visits of both types of vehicle. Let $V^c_s$ be the set of vertices which are the copies of special customers $V_s$. Let $V_1$ be defined as $V_1 = V_r \cup V^c_s$. It is considered as the set of customer vertices that must be visited by regular vehicles. Each vertex $i$ in $V_1$ is associated with a demand $q_i$. Define $v^s_j$ and $v^e_j$ ($j = \{1, 2\}$) be the vertices at which the vehicle of type $j$ starts from and comes back to, respectively. Here, $j = 1$ represents regular vehicles and $j = 2$ is the index for special vehicles. Note that, the vertices $v^s_j$ and $v^e_j$ ($i = \{1, 2\}$) share the same location as $v_0$. Additionally, we define $V^+_1 = V_1 \cup \{v^s_1\} \cup \{v^e_1\}$ be the set of vertices appearing on routes of regular vehicles; and $V^+_2 = V_s \cup \{v^s_2\} \cup \{v^e_2\}$ be the set of vertices visited by special vehicles. A service time $s^1_i$ is associated with a vertex $V^+_1$, and a service time $s^2_i$ is with a vertex $i \in V^+_2$. Note that the service times at the depot and its copies are set to null. A time window ($l_i, u_i$) is imposed on each vertex $i \in V^+_1 \setminus \{v^s_1\}$. Finally, each edge $(i, j) \in E$ is associated with non-negative values $c^k_{ij}$ and $t^k_{ij}$ representing the travel cost and travel time between vertices $i$ and $j$ of vehicle type $k$. 

The problem then consists in constructing routes for the fleet of vehicles such that the total travel cost incurred by the fleet of vehicles is minimized and the following constraints are satisfied:
\begin{itemize}
\item Each vehicle must begin the route at the depot, deliver services to customers and finally return to the depot.
\item Each regular customer is served by exactly one regular vehicle.
\item Each special customer is served by exactly one regular vehicle and one special vehicle.
\item The total demand serviced by a regular vehicle must not exceed its capacity $Q$.
\item A regular vehicle must start its service at a vertex $i \in V^+_1 \setminus \{v^s_1\}$ within the time window $(l_i, u_i)$.
\item Starting service time at a special customer $i \in V_s$ visited by a special vehicle must be within a time window $[t_{r_i}-\alpha_i, t_{r_i}+\beta_i]$. Here, $\alpha_i$ and $\beta_i$ are given parameters representing a possible delay between regular and special services at vertex $i$. And, $t_{r_i}$ is the starting service time at vertex $r_i$ (the copy of $i$ in regular vehicle route). 
\end{itemize}

 We now present the new CP model for the problem. First, it is worth mentioning that unlike MIP formulations, there is no standard in CP formulation because it strongly depends on each CP package. In this study, we formulate the model using generic keywords and syntaxes of IBM ILOG CP Optimizier \cite{CPO} adapted from the CP formulations proposed by \cite{Laborie2009, Ghedira2013, Goel2015, Ham2018}. Our model uses the following variables:

\subsection*{Variables}
\begin{tabularx}{\textwidth}{llX}
$Itv^r_{i}$   &         & interval variable that represents the time interval of size $s^1_i$ vertex $i \in V^+_{1}$ is visited\\  
$ItvAlt^r_{ij}$   &         & optional interval variable that represents the time interval vehicle $j \in M_{1}$ visits vertex $i \in V^+_{1}$;\\
$Seq^r_{j}$   &         & sequence variable that represents all working time intervals of vehicle $j \in M_{1}$;\\

$Itv^s_{i}$   &         & interval variable that represents the time interval of size $s^2_i$  vertex $i \in V^+_{2}$ is visited;\\
$ItvAlt^s_{ij}$   &         & optional interval variable that represents the time interval vehicle $j \in M_{2}$ visits customer $i \in V^+_{2}$;\\
$Seq^s_{j}$   &         & sequence variable that represents all working time intervals of vehicle $j \in M_{2}$;\\
\end{tabularx}

An interval variable represents the interval of time during which a task can occur. It contains a starting point, an end point, a size, and it can be optional. A decision variable is used to represent whether or not an interval is present. If an interval is marked as optional, it may be absent in the solution. Sequence is a type of variable in IBM ILOG OPL which can be empty or can contain a subset of variables. A sequence represents all intervals that are present in the solution. The constraints in our model are as follows:

\subsection*{Constraints}
\begin{enumerate}

\item Function to link each interval to a location.
\begin{align}
 \textit{\textbf{type}} \; \text{function} \; \theta (Seq^r_{j}, ItvAlt^r_{ij}) =  i \qquad\text{}\qquad \forall i \in V^+_{1}, j \in M_{1} \\
\textit{\textbf{type}} \; \text{function} \;\theta (Seq^s_{j}, ItvAlt^s_{ij}) =  i \qquad\text{}\qquad \forall i \in V^+_{2}, j \in M_{2}
\end{align}
	
A $\theta$ value is defined for each pair of ($Seq^r_{j}$, $ItvAlt^r_{ij}$) and ($Seq^s_{j}$, $ItvAlt^s_{ij}$). The type $\theta$ of the interval variable is its last known location. 

\item Each customer must be served by one vehicle of the corresponding type.
\begin{align}
	\textit{\textbf{alternative}} \; ( Itv^r_{i}, ItvAlt^r_{ij}: j \in M_{1} ) \qquad\text{}\qquad \forall i \in V^+_{1}\\
	\textit{\textbf{alternative}} \; ( Itv^s_{i}, ItvAlt^s_{ij}: j \in M_{2} ) \qquad\text{}\qquad \forall i \in V^+_{2}
\end{align}

\textbf{\textit{alternative}} function ensures that exactly one set of intervals $ItvAlt^r_{ij}$ (or $ItvAlt^s_{ij}$) is present in the solution; and interval variable starts and ends together with the interval variable $Itv^r_{i}$ (or $Itv^s_{i}$).

\item Travel time between two customers must be taken into account. In the following, $T_k = \{t^k_{ij}\}$ is the matrix representing travel times between two vertices $i$ and $j$ in the set $V^+_{k}$.
\begin{align}
\textit{\textbf{noOverlap}} \;( Seq^r_{j}, T_1) \qquad\text{}\qquad \forall j \in M_{1}\\
\textit{\textbf{noOverlap}} \; ( Seq^s_{j}, T_2) \qquad\text{}\qquad \forall j \in M_{2}
\end{align}

\textit{\textbf{noOverlap}} constraint on sequence variables $Seq^r_{j}$ and $Seq^s_{j}$ states that the sequence defines a chain of non-overlapping intervals, and any interval in the chain is constrained to end before the start of the next interval in the chain. 

\item All vehicles start their route at the starting depot.
\begin{align}
\textit{\textbf{first}} \; ( Seq^r_{j}, ItvAlt^r_{v^s_{1}j})\qquad\text{}\qquad \forall j \in M_{1}\\
\textit{\textbf{ first}} \; ( Seq^s{j}, ItvAlt^s_{v^s_{2}j})\qquad\text{}\qquad \forall j \in M_{2}
\end{align}

\textit{\textbf{first}}$(p, j)$ function states that if interval $j$ is present, it will be the first interval in the sequence $p$. These two constraints force all regular vehicles to start their routes at $v^s_{1}$ and all special vehicles to start at $v^s_{2}$.

\item All vehicles finish their routes at the corresponding ending depot.
\begin{align}
\textit{\textbf{last}} \; ( Seq^r_{j}, ItvAlt^r_{v^e_{1}j}) \qquad\text{}\qquad \forall j \in M_{1}\\
\textit{\textbf{last}} \; ( Seq^s_{j}, ItvAlt^s_{v^e_{2}j}) \qquad\text{}\qquad \forall j \in M_{2}
\end{align}

Similar to \textit{\textbf{first}}($p, j$) function, \textit{\textbf{last}}($p, j$) function states that if interval $j$ is present, it will be the last interval in the sequence $p$. These two constraints are to force all regular vehicles finishing their routes at ending depot $v^e_{1}$ and all special vehicles finishing their routes at ending depot $v^e_{2}$.

\item Time window constraints.
\begin{align}
	 l_{i} \leq \textit{\textbf{startOf}}(Itv^r_{i}) \leq u_{i}\qquad\text{}\qquad \forall i \in V_{1} \cup \{v^e_1\}
 \end{align}

\textit{\textbf{startOf}}$(j)$ represents the start of interval $j$ whenever the interval variable $j$ is present. 

\item Capacity constraints.
\begin{align}
 \sum_{i \in V_{1}}^{}q_i.  \textit{\textbf{presenceOf}} \; (ItvAlt^r_{ij}) \leq Q \qquad\text{}\qquad \forall j \in M_{1}
\end{align}

\textit{\textbf{presenceOf}}$(j)$ is equal to 1 if interval variable $j$ is present in the solution, 0 otherwise.

\item Synchronization constraints.
\begin{align}
	\textit{\textbf{startOf}}\;(Itv^r_{r_{i}}) - \alpha_{i} &\leq \textit{\textbf{startOf}}\;(Itv^s_{i}) \qquad \forall i \in V_{s}\\
    \textit{\textbf{startOf}}\;(Itv^s_{i}) &\leq \textit{\textbf{startOf}} \; (Itv^r_{r_{i}}) + \beta_{i} \qquad \forall i \in V_{s}
\end{align}

\end{enumerate}

\subsection*{Objective function}
\paragraph\
We compute the total cost traveled by regular and special vehicles as follows:\\
 $Cost^r_{j}$  = $\sum_{i \in V^+_{1}} c^1_{ik} \; \forall j \in M_1$ where\\
\qquad\text{}\qquad$k$ = \textit{\textbf{typeOfNext}} $(Seq^r_{j}, ItvAlt^r_{ij}, i, i)$.\\  $Cost^s_{j}$  = $\sum_{i \in V^+_{2}} c^2_{ik} \; \forall j \in M_2$ where\\
\qquad\text{}\qquad$k$ = \textit{\textbf{typeOfNext}} $( Seq^s_{j}, ItvAlt^s_{ij}, i, i)$.\\

Then the objective function can be written as:
\begin{align}
\textbf{Minimize} \; \sum_{i \in M_{1}} Cost^r_{i} + \sum_{i \in M_{2}} Cost^s_i
\end{align}

\section{Linear programming-based adaptive large neighborhood search algorithm}
\label{method}
% In general, the structure of our ALNS is described as follows. Starting from an initial solution, the algorithm iteratively choose a destroy and a repair operator based on their performance given by scores and weights from previous segment of iterations. With a chosen pair of operators, some customers are first removed from the solution using the destroy operator. Then, using the according repair operator customers are re-inserted into the routes. In case the resulting solution $s^\prime$ meets the acceptance criteria it will replace the current solution $s$. In case $s^\prime$ is better than the best solution so far, it replaces {$s_{best}$}. The scores and weights of the destroy and repair operators are updated after each segment of 100 iterations. This process keeps going on until some stopping criterion is met. 

To tackle the VRPSC problem, we design an Adaptive Large Neighborhood Search (ALNS) heuristic, which is based on a Large Neighborhood Search (LNS) introduced by \cite{Shaw}. At each iteration, LNS explores a large neighborhood, which can rearrange a large part of the current solution, therefore allowing the search to move to other promising search spaces.

More precisely, LNS	decomposes the original problem	by unfixing some decision variables, leading to a partial	solution. The unfixed decision variables define	a neighborhood of solutions that can be explored by a specific procedure via, possibly, a heuristic or a Mixed Integer	Programming	(MIP) solver. If the procedure finds	an improved solution, it becomes the new current solution and a new large neighborhood is defined around it. This process is repeated until a stopping criterion is reached. 

A first	key point is the selection of fixed variables to create a partial solution. In fact, the number of fixed variables impacts the size of the neighborhood (the more fixed variables, the narrower the neighborhood). A common	strategy is	to dynamically vary	the number of removed variables. A second key point lies in the selection of fixed/removed variables which can use a random choice or a	more sophisticated strategy to guide the search. Finally, the	procedure that explores	the neighborhood should provide	good quality solutions in a short amount of time. Adaptive Large Neighborhood Search is an extension	of	LNS	with a number of different insertion and removal operators.	In	comparison with	LNS, a component that adaptively chooses among a set of removal and insertion	heuristics is added to the algorithm. The pseudo-code of a ALNS to solve problems with minimizing objective function is shown in Algorithm \ref{alg:alns}. At each iteration,	a randomly selected pair of operators (with procedures	SelectDestruction and SelectRepair, lines 5 and 6) is	applied	to	the	current	solution (line	7),	with	probabilities (respectively, sets $p^{remove}$, $p^{insert}$ ) updated	by	a learning	process	(line	12). The more an operator $i$ has contributed to the solution process, the larger probability $p_i$ it has of being chosen.

\begin{algorithm}[h] \label{alg:alns}
\caption{General ALNS}
% \SetAlgoNoLine
% \SetAlgoNoEnd
\DontPrintSemicolon
\SetKwFunction{FALNS}{ALNS}
Create an initial solution s\;
$s_{best} := s$\;
Initialize $p^{remove}$ and $p^{insert}$\;
\While{the stop-criterion is not met} {
    $re$ := SelectRemoval($p^{remove}$)\;
    $in$ := SelectInsertion($p^{insert}$)\;
    solution $s'$ = GenerateNewSolution($s,re,in$)\;
    \If{cost($s'$) $<$ cost($s_{best}$)}{$s_{best} := s'$} 
    \If{accept(s', s)}{$s := s'$}
    update $p^{remove}$ and $p^{insert}$\;
}
\KwRet $s_{best}$ \;

\end{algorithm}

% In lines 5 and 6, a number of requests are removed from the current solution and then reinserted to the solution again by using single remove and insert method. For ALNS heuristic, instead of using single method for remove and insert, we used combine of different insert and remove heuristic. In each iteration, we select a pair of heuristic based on an adaptive mechanism.

% The accept criteria in line 10 determines if new solution $s'$ should be accept. In LNS implementations by Shaw (1998), all improving solution would be accepted. However, only accepting improving solution could make the heuristic get trapped at local optimum. In this paper we use a Simulated Annealing(SA) accept criteria.

% The heuristic will stop after meeting some stop condition. In our implementation, the algorithm stop after a fixed number of iteration.

\subsection{Insertion operators}
\subsubsection{Cheapest insertion heuristic}
The purpose of the insertion operation is to reinsert unserviced requests into solution. For this task, one can use the cheapest insertion heuristic which inserts the customer into a route at a feasible position making the objective value increase the least. The process is repeated until all customers are serviced or no more customers can be inserted. The insertion cost of a regular customer $k$ into a regular route positioned between two consecutive vertices $i$ and $i+1$ (denoted by $IC^1_k$) is computed as:

\begin{align}
    IC^1_k = c^1_{ik} + c^1_{k(i+1)} - c^1_{i(i+1)}
\end{align}

Whenever a special customer is considered for insertion, it will be added to two positions: one on regular routes and another on special routes. This must be incorporated when computing the insertion cost of special customers. The average value is used to compute the insertion cost of a special customer $k$ at positions between vertices $i$ and $i+1$ on regular routes and between vertices $i'$ and $i'+1$ on special routes as follows:
\begin{align}
   IC^2_k = \frac{(c^1_{ik} + c^1_{k(i+1)} - c^1_{i(i+1)}) + (c^2_{i'k} + c^2_{k(i'+1)} - c^2_{i'(i'+1)})}{2}
\end{align}

\subsubsection{Regret heuristics}
As in \cite{Ropke}, we also use regret-$k$ heuristics as repair operators. Instead of selecting the customer with the least insertion cost in each construction step, the regret heuristics select the customer with the highest regret-$k$ value, computed as follows: we denote $f_{i,j}$ is the insertion cost when inserting customer $i$ into the best position of route $j$. If this insertion is infeasible w.r.t time window and synchronization constraints, the insertion cost is set to infinity, i.e. $f_{i,j} = \infty$. Let $r_{ik}$ be the route on which vertex $i$ has the $k$-th lowest insertion cost. The regret-$k$ value $RV_i$ of customer $i$ is then calculated as:

\begin{align}
RV_i = \sum_{j=1}^{k} (f_{(i,r_{ij})} - f_{(i,r_{i1})})
\end{align}

The regret-$k$ heuristics choose the unvisited customer $i$ with the highest regret-$k$ value $RV_i$ and insert it into the feasible position leading the least insertion cost. Ties are broken by selecting customers with the lowest insertion cost $f_{(i,r_{i1})}$. Informally speaking, we choose the insertion that leads to the most regret, if it is not done at present. In some situations, if a vertex can be inserted somewhere in the current solution, but cannot be inserted in at least $k$ routes, then the vertex that can be inserted in the fewest number of routes is selected. This ensures that the vertex which does not have many insertion options in the current solution will be considered first.

It can be observed that the cheapest insertion heuristic, which we mentioned earlier, is the special case of the regret heuristic with $k = 1$ due to the tie-breaking rule. For any $k>1$, the regret heuristic looks further into future solutions to decide the choice of insertion. In this research, we use regret heuristics with $k = \{2, 3\}$ to design insertion operators of our ANLS.

\subsubsection{Checking insertion feasibility of regular customers}

When inserting a vertex into a position of the current partial solution, it is required to verify if the insertion satisfies the capacity, time window and synchronization constraints. As the insertion operation is repeated multiple times during the search, designing a quick verification procedure is critical to speed up the overall algorithm. As the capacity constraint is easily checked in $\mathcal{O}(1)$, we focus on the time window and synchronization constraints only. Verifying the feasibility of an insertion operation w.r.t these constraints are more complex because they delay  subsequent visits leading to other violations. As proposed in \cite{Kindervater}, the time window constraint can be checked in $\mathcal{O}(1)$ by pre-computing the maximum delay (push forward) that is allowed at each arc of the current solution, without violating time windows. In this research, we also reuse this idea to handle both time window and synchronization constraints when inserting regular vertices.  

Given a partial solution, in order to consider all  possible positions to insert an unserved regular customer $i$ into the routes, we calculate the maximum duration of time (also called maximum delay) that can be spared after a vehicle finishes serving vertex $p-1$ and before it starts to serve next vertex $p$ without violating any constraints of other nodes. This value is denoted as $\delta_{a}$, where $a$ represents the arc from $p-1$ to $p$. We calculate the maximum delays at all the arcs of the current solution using the following linear programming model:

Let $\overline{V_1}$ and $\overline{V_2}$ be the set of all vertices visited by regular and special vehicles in current solution $s$, respectively. Denote $\overline{A} = A_1 \cup A_2$ is the set of arcs forming the routes in $s$; $A_1$ and $A_2$ are the sets of arcs on regular and special routes, respectively. We use two types of variables: $\tau_{i}$ implying starting service time at customer $i$ and $\delta_a$ representing the maximum delay on arc $a = (p-1, p)$.

\begin{align}
\text{(F1)}  \quad \quad \textbf{Maximize} \qquad \delta_a \label{obj1}\\
\textbf{Subject to} \qquad l_{i} \leq \tau_{i} &\leq u_{i}\quad\quad\quad \forall i \in \overline{V_1} \label{mod1-1}\\
-\alpha_i \leq \tau_{i}-\tau_{r_i} &\leq \beta_i \quad\quad\quad \forall i \in \overline{V_2} \label{mod1-2}\\
\tau_{i-1} + s^k_{i-1} + t^k_{(i-1, i)}  &\leq \tau_{i} \qquad \forall (i-1, i) \in A_k \setminus \{a\}, k \in \{1, 2\} \label{mod1-3}\\
\tau_{p-1} + s^1_{p-1} + t^1_{(p-1, p)} + \delta_{(p-1, p)} &\leq \tau_{p}  \label{mod1-4} \\
\tau_i &\geq 0 \qquad \forall i \in \overline{V_1} \cup \overline{V_2} \label{mod1-5}\\
\delta_a &\geq 0  \label{mod1-6}
\end{align}

Objective (\ref{obj1}) is to maximize the maximum delay on arc $a$. Constraints (\ref{mod1-1}) and (\ref{mod1-2}) respectively ensure time window and synchronization constraints at all vertices of the current solution. Constraints (\ref{mod1-3}) represent the relationship between the starting times at vertices $(i-1)$ and $i$ when a vehicle travels from $(i-1)$ to $i$. Constraint (\ref{mod1-4}) has the same meaning as constraints (\ref{mod1-3}) but is written for arc $a$. Finally, constraints (\ref{mod1-5}) and (\ref{mod1-6}) define the domain of variables.  

After all the maximum delays are available, we check if an unserved regular node $i$ can be inserted at the position between node $p-1$ and node $p$ by computing the earliest arrival time ($arrivalTime$) and the waiting time ($waitTime$) at $i$ as follows: 
\begin{align} 
 &arrivalTime_i = \tau_{p-1} + s^1_{p-1} + t^1_{(p-1)i} \nonumber\\
&waitTime_i = \max(l_i - arrivalTime_i, 0) \nonumber
\end{align}

Finally, we check if feasibility of the insertion satisfies the following constraints: 
\begin{align}
t^1_{(p-1)i} + t^1_{ip} - t^1_{(p-1)p} + waitTime_i + s^1_i  &\leq \delta_{(p-1, p)} \label{insertCost}\\
arrivalTime_i &\leq u_i \label{twOfi}
\end{align}

Constraint (\ref{insertCost}) ensures that the insertion does not lead to a violation of time window and synchronization constraints at all the customers in the current solution. Constraint (\ref{twOfi}) verifies the time window constraint of vertex $i$.

 Although it is fast to solve a LP model of type (F1), the running time of the insertion operators is still expensive due to the large quantity of LP models solved during the search. Through observation, we note that constructing the model to find maximum delay on each arc takes more computational time than solving it. As such, we propose the following model to reduce the running time of the insertion operators: 

\begin{align}
\text{(F2)}  \quad \quad \textbf{Maximize} \qquad \sum_{i \in \overline{A}} \varpi_i\delta_i \label{obj2}\\
\textbf{Subject to} \qquad (\ref{mod1-1}), (\ref{mod1-2}), (\ref{mod1-5}) \nonumber\\
\tau_{i-1} + s^k_{i-1} + t^k_{(i-1, i)} + \delta_{(i-1, i)} &\leq \tau_{i} \qquad \forall (i-1, i) \in A_k, k \in \{1, 2\} \label{mod2-1}\\
\delta_i &\geq 0 \qquad \forall i \in \overline{A} \label{mod2-2}
\end{align}

Objective (\ref{obj2}) is to maximize the weighted maximum delay at all arcs of the solution. Here, $\varpi_i$ is a given binary coefficient representing the weight of arc $i \in \overline{A}$. Constraints (\ref{mod2-1}) indicate the relationship between the variables $\delta$ and $\tau$. To find the maximum delay on an arc $i$, we just need to set its weight $\varpi_i$ to 1 and weights of other arcs to null. The proposed model (F2) allows us to save a lot of time by constructing a model once and then creating a new one by changing two coefficients in the objective function only. As a result, we can avoid constructing multiple models from scratch. A preliminary experiment shows that using this method helps reduce at least 30\% running time of the overall algorithm. After each model calculating the maximum delay at an arc with two extremities $p-1$ to $p$ is solved, the value of the variables $\tau_{p-1}$, $\tau_p$, and $\delta_{(p-1, p)}$ are also saved for checking insertion feasibility of special customers described in the following section.

\subsubsection{Checking insertion feasibility of special customers}

Whenever a special customer is selected to be added into the current solution, it will be inserted into two positions: one on regular routes and the other on special routes. Multiple insertion operations make it impossible to use the maximum delay for the feasibility verification purpose. The following LP model, without objective function (F3), is used to check if a special vertex $j$ can be added in arc $a_1 = (p_1-1, p_1)$ on a regular route and arc $a_2 = (p_2-1, p_2)$ on a special route:

\begin{align}
\text{(F3)} \quad \textbf{Subject to} \qquad l_{i} \leq \tau_{i} &\leq u_{i}\quad\quad\quad \forall i \in \overline{V_1} \cup \{r_j\} \label{mod3-1} \\
-\alpha_i \leq \tau_{i}-\tau_{r_i} &\leq \beta_i \quad\quad\quad \forall i \in \overline{V_2} \cup \{j\} \label{mod3-2}\\
\tau_{i-1} + s^k_{i-1} + t^k_{(i-1, i)}  &\leq \tau_{i} \qquad \forall (i-1, i) \in A_k \setminus \{a_k\}, k \in \{1, 2\} \label{mod3-3}\\
\tau_{p_1-1} + s^1_{p_1-1} + t^1_{(p_1-1, r_j)}  &\leq \tau_{r_j}  \label{mod3-4} \\
\tau_{r_j} + s^1_{r_j} + t^1_{(r_j, p_1)}  &\leq \tau_{p_1}  \label{mod3-5} \\
\tau_{p_2-1} + s^2_{p_2-1} + t^2_{(p_2-1,j)}  &\leq \tau_{j}  \label{mod3-6} \\
\tau_{j} + s^2_{j} + t^2_{(j, p_2)}  &\leq \tau_{p_2}  \label{mod3-7} \\
\tau_i &\geq 0 \qquad \forall i \in \overline{V_1} \cup \overline{V_2} \cup \{j, r_j\} \label{mod3-8}
\end{align}

The meaning of variables $\tau$, other parameters, and constraints (\ref{mod3-1})-(\ref{mod3-3}) can be derived from (F1) and (F2). Constraints (\ref{mod3-4})-(\ref{mod3-7}) are similar to (\ref{mod3-3}), but written for 4 new arcs created by the insertion of vertices $j$ and its mirror $r_j$: ($p_1 -1, r_j$), ($r_j, p_1$), ($p_2 -1, j$), and ($j, p_2$). 

As mentioned above, after solving each model computing the maximum delay on an arc ($p-1, p$), the starting service times of nodes $p-1$ and $p$, or in other words, the values of $\tau_{p-1}$ and $\tau_p$ are saved. $\tau_{p-1}$ can be seen as the earliest time the vertex $p-1$ can be serviced while $\tau_p$ can be seen as the latest time vertex $p$ can be serviced. To avoid misunderstanding these notations, we denote $\tau_{p-1}$ as $et_{p-1}$ and $\tau_p$ as $lt_p$. Using these saved values, we can efficiently check if special node $i$ and its copy $r_i$ can be inserted on arc $a_2 = (p_2-1, p_2)$ of a special route and arc $a_1 = (p_1 - 1, p_1)$ of a regular route, respectively. First, lower bound (denoted by $lb$) and upper bound (denoted by $ub$) of arrival times at $r_i$ when being inserted on arc $a_1$ and $i$ when being inserted on arc $a_2$ are computed as follows:
\begin{flushleft}
$lb_{r_i}$= $et_{p_1-1} + s^1_{p_1-1} + t^1_{p_1-1,r_i}$\\
$ub_{r_i}$= $lt_{p_1} - s^1_{r_i} - t^1_{r_i,p_1}$\\
$lb_i$= $et_{i} + s^2_{p_2-1} + t^2_{p_2-1,i}$\\
$ub_i$= $lt_{p_2} - s^2_{i} - t^2_{i, p_2}$.\\
\end{flushleft}

It can be seen that the synchronization constraints will be violated in the following two cases: 
\begin{align}
lb_i-ub_{r_i} > \beta_i \qquad \text{or} \qquad ub_{i}-lb_{r_i} < - \alpha_i \label{condi1}
\end{align}

The insertions of $i$ and $r_i$ also need to satisfy the time window constraint at node $r_i$ and the maximum delays computed from the model F2. Thus, we can utilize this property to rapidly verify the feasibility of the insertions. Unlike regular vertices, possible waiting times at $i$ and $r_i$ are created by not only time window constraints, but also synchronization constraints. The lower bounds of waiting times created by time window ($waitTime^{tw}_j$) and synchronization ($waitTime^{sync}_j$) at a node $j$ can be computed as follows:
\begin{flushleft}
$waitTime^{tw}_{r_i}$ = max(0, $l_{r_i} - lb_{r_i}$)\\
$waitTime^{tw}_{i}$ = 0\\
$waitTime^{sync}_{r_i}$ = max(0, $lb_{i} - \beta_i - lb_{r_i}$)\\
$waitTime^{sync}_{i}$ = max(0, $lb_{r_i} - \alpha_i - lb_i$)\\
\end{flushleft}

\begin{figure}[H]
  \centering
  \includegraphics[scale = 0.5]{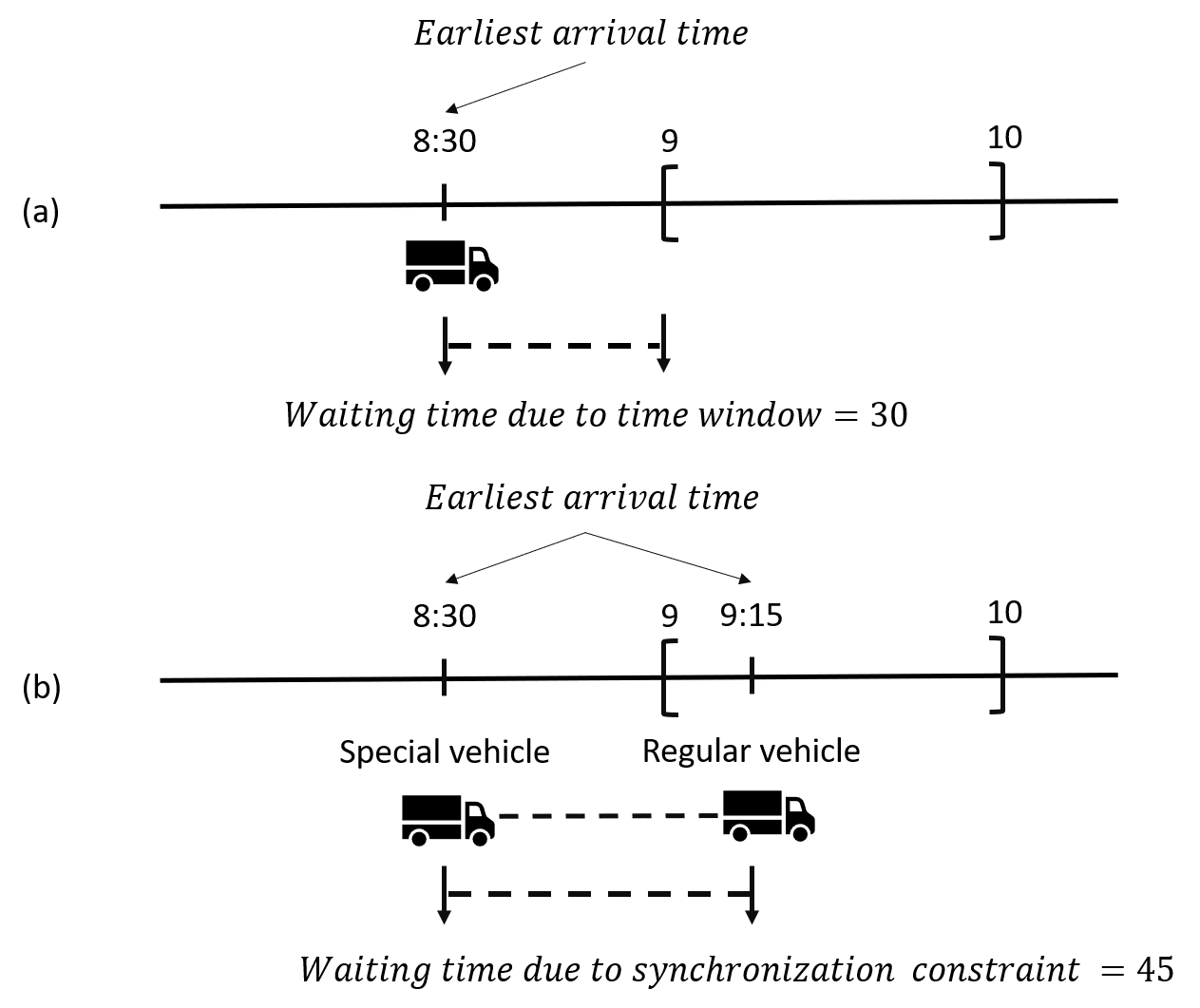}
  \caption{Computing waiting times due to time window and synchronization constraints}
  \label{fig:example}
\end{figure}

Figure \ref{fig:example} illustrates the computation of waiting times in case of $\alpha = 0$ and $\beta = 10$. In Figure 1a, the vehicle arrives at the customer location before the time window and has to wait 30 minutes before starting delivery. In Figure 1b, the special vehicle arrives before the regular vehicle, but it has to wait until the regular vehicle starts to service the customer because the value of $\alpha$ is zero. Thus, the waiting time due to the synchronization constraint, in this case, is 45 minutes.

Hence, we can calculate the lower bounds of waiting times at node $j$ by taking the maximum value between $waitTime^{tw}_j$ and $waitTime^{sync}_j$:
\begin{flushleft}
$waitTime_{r_i}$ = max($waitTime^{tw}_{r_i}$, $waitTime^{sync}_{r_i}$)\\
$waitTime_{i}$ = max($waitTime^{tw}_{i}$, $waitTime^{sync}_{i}$)
\end{flushleft}

After all the values above are calculated, we can skip the insertions which violate one of the following constraints: 
\begin{align}
t^1_{(p_1-1)r_i} + t^1_{r_ip_1} - t^1_{(p_1-1)p_1} + waitTime_{r_i} + s^1_{r_i} &\leq \delta_{(p_1-1, p_1)} \label{condi2}\\
t^2_{(p_2-1)i} + t^2_{ip_2} - t^2_{(p_2-1)p_2} + waitTime_{i} + s^2_{i}  &\leq \delta_{(p_2-1, p_2)} \label{condi3}\\
lb_{r_i} &\leq u_{r_i} \label{condi4}
\end{align}
The constraints (\ref{condi2}) and (\ref{condi3}) verify if the insertions satisfy the maximum delays while constraint (\ref{condi4}) checks the time window at the regular vertex $r_i$. In addition, based on the characteristic of the $k$-regret heuristics, we can only consider the insertions if their cost is smaller than the current $k$-th best insertion cost.

It is worth mentioning that validation procedures (\ref{condi1})-(\ref{condi4}), which run in $\mathcal{O}(1)$ if the values of the variables of each program F2 are available, can detect plenty of infeasible insertions, thus saving a lot of run time of the algorithm. A preliminary experiment on instances with 50 customers and 25 synchronizations shows that our fast validation procedures help to increase the algorithm's speed by up to 20 times. This ratio increases on larger instances with more synchronizations. However, if the insertion passes all the validations, we cannot ensure that the insertion is indeed feasible w.r.t the time window and synchronization constraints. As a consequence, whenever an insertion of a special request passes the checks above, we will continue to examine if that insertion is feasible by solving a program of type (F3). If that model returns a solution, then the insertion is feasible, and infeasible otherwise.
%In order to minimize the number of feasibility linear programming model that we must solve, when finding the best position in each route to insert a request, if the insertion cost is larger than the current minimum insertion cost, we will discard the check at that position since it is unnecessary to inspect insertion at this position according to regret heuristic.

\subsection{Removal operators}
The destroy operators remove a fraction of vertices from a complete solution based on different criteria, each guiding the algorithm to another search space. The input of the operators is a complete solution $s$ and their outputs are $nb_{rm}$ vertices that have been removed from the $s$. We use three destroy operators originally proposed by \cite{Ropke}: random removal, related removal and worst removal. 

\subsubsection{Random Removal}

This is the simplest removal operator. It randomly selects $nb_{rm}$ vertices in the solution and removes them. Other vertices remain unchanged. This obviously helps the algorithm diversify the search.

\subsubsection{Related Removal}
The idea of the related removal, as its name indicates, is to remove similar vertices with the expectation that they could interchange their positions to create a better solution. More specifically, to measure the similarity between two vertices $i$ and $j$, we use the \textit{relatedness} $R_{ij}$ which is calculated as follows: 
\begin{align}
    R_{ij} = \lambda_1\frac{|\tau_{i} - \tau_{j}|}{maxTime} + \lambda_2\frac{d_{ij}}{maxDis} + \lambda_3\frac{|q_i - q_j|}{maxDem} + \lambda_4 |type_i - type_j|
\end{align}

To calculate the relatedness of two vertices, we take into account the differences of four characteristics: their starting service times ($\tau$); the distance between them ($d_{ij}$); their demand size; and their type. Note that the value of $\tau$ is obtained from solving a program of type F3 whenever a new complete solution is found; and $type_i$ is set to 1 if vertex $i$ is special, and to 0 if it is regular. The difference is normalized such that it only takes values from the interval [0, 1]. In this formula, $maxTime$, $maxDis$, and $maxDem$ indicate the largest starting service time, the largest distance, and the largest demand of all vertices in the solution, respectively. In addition, each characteristic $i$ is associated with a weight $\lambda_i$ to measure its importance. 
\subsubsection{Worst Removal}
The worst removal operator removes the vertices that are very expensive, with the expectation that these vertices might be located in wrong places. Given a request $i$ served by some vehicle in a solution $s$, we define the cost of the vertex $\Delta_i$ as the difference between the cost of $s$ and the cost of the new solution where vertex $i$ is removed completely from $s$. The worst removal heuristic repeatedly chooses a vertex $i$ with the largest cost $\Delta_i$ until $nb_{rm}$ vertices have been removed. 

To add more diversification to our algorithm, the related and worst removal operators are randomized by removing the $\lfloor{y^{p_r}|R|}\rfloor$-th most related (or expensive) request where $R$ is the set of vertices in solution and $y$ is a random number in [0, 1], and parameter $p_r$ is used to control the randomization. If $p_r$ is small, the most related (in the case of related removal) or expensive (in the case of worst removal) vertex is selected, while less related (or expensive) vertices may be chosen for larger values of $p_r$ with a probability that decreases with the cost $\Delta_i$. The values of $p_r$ are taken from \cite{Ropke}.

Finally, our lp-ALNS also uses acceptance criteria embedded in a simulated annealing framework, adaptive score adjustment to select operators in a dynamic fashion, and adding noise to insertion cost to increase the diversification. All these components and their parameter settings are taken from \cite{Ropke} without any change.

\section{Computational results}
\label{result}
In this section, the effectiveness of the proposed algorithms is examined. We test our algorithms on the instances proposed in \cite{Hojabri} with the number of customer $|V|$ = 25, 50, 100, and 200. These instances are generated from the VRPTW instances of \cite{Solomon1987, Homberger1999} containing three types, depending on the customers' distribution. The customers are randomly located in the instances of type R, clustered in type C, and mixed between randomly located and clustered in type RC. The instances are also categorized into two classes based on the capacity of vehicles. The first class (including C1, R1, RC1) consists of instances with a relatively small capacity $Q$ compared to the total customer demand, while in the second class (C2, R2, RC2), the capacity is relatively large. Note that in the instances of types R and RC, the vertices are identically distributed in class 1 and 2, while this is not true for type C. And finally, in these VRPTW instances, the travel time and travel cost between two vertices are set to their Euclidean distance.

The original VRPTW instances are transformed into VRPSC instances as follows: the number of special customers $|V_s|$ is set to $\lceil n_s.|V| \rceil$ where $n_s$ is the percentage of special customers. There are three values for $n_s$: 5\%, 25\%, and 50\%. More precisely, the first customer in the VRPTW instances is considered a special customer and the next special customers are selected using a constant interval defined by $\frac{1}{n_s}$. In the case of the synchronization constraint, the values of $\alpha_i$ and $\beta_i$ are set to 0 and 10 for every special customer $i \in V_s$, respectively. And finally, by private contact, it turns out that the authors in \cite{Hojabri} report inexact results for three instances C101, C105 and C106 with 25 customers and 2 synchronizations, so we removed these instances from our experiments.

The CP model is coded in IBM OPL 12.8.0 while the lp-ANLS is implemented in C++ using CPLEX 12.8.0 for the resolution of the linear programs. Both methods are run on a 3.20GHz Xeon(R) E5-2667. Note that our reference algorithm (the cp-ALNS of \cite{Hojabri}) was run on a 3.07GHz Xeon(R) X5675, which is a similar generation to our processor. Since different CPU speed conversion techniques can provide very different results, we decided to present the raw running time, letting the readers choose their preferred approach. The parameter setting of the lp-ALNS is chosen empirically. We have tested many settings and the following setting gives the best performance, in terms of both quality and computational time for our algorithm. The number of removed vertices in the removal operators $nb_{rm}$ is a random integer between 4 and $\min(40, \lfloor 0.4*|V| \rfloor)$. In related removal operator, the values of $\lambda_1$-$\lambda_4$ are set to 4, 2, 1, and 4, respectively. The lp-ALNS stops after 25000 iterations. All the detailed results can be found in \textit{http://www.orlab.vn/home/download}.

In the first experiment, we compare the results obtained by our CP model and lp-ALNS with those of cp-ALNS proposed by \cite{Hojabri}. Because our CP model cannot handle the large instances, we chose the small instances with 25 customers for the experiment. The limited running time of the CP approach for each instance is set to 5 minutes and 3 hours. Table \ref{tab1} shows the number of times each of our methods finds better solutions (Columns ``Better"), equal solutions (Columns ``Equal"), and worse solutions (Columns ``Worse") compared to cp-ALNS. The columns ``Gap" report the average gaps (in percentage) between solution costs of our methods and those of cp-ALNS. The negative values in these columns indicate that our methods provide better solutions in terms of objective function values. The  results obtained show that although our CP model-based algorithms cannot solve any instance to optimality in 3 hours, they do provide quite good solutions. Remarkably, the CP model performs better than cp-ALNS on 56 instances in a much shorter running time (5 minutes vs a couple of hours of cp-ALNS). It can be observed that CP models work better on the instances of the first class (C1, R1, and RC1) and worse on the instances of second class. The instances with shorter routes tend to be easier for our CP model. The results clearly show the performance of our lp-ALNS. It is the most efficient method in terms of solution quality, as it provides 128 better solutions compared to cp-ALNS and is worse on only 7 over 165 instances. Moreover, the gaps, on average, are negative on all instance classes (except RC12). Our lp-ALNS can averagely improve the objective values up to 4.59\% (instance class R1).

% \usepackage{multirow}
% \usepackage{multirow}

\begin{table}[H]
\centering
\setlength{\tabcolsep}{1.5pt}
\begin{tabular}{|c|c|c|c|c|l|c|c|c|l|c|c|c|l|} 
\hline
\multirow{2}{*}{Data} & \multirow{2}{*}{Sync} & \multicolumn{4}{c|}{lp-ALNS}     & \multicolumn{4}{c|}{CP (5 min)}  & \multicolumn{4}{c|}{CP (3h)}      \\ 
\cline{3-14}
                      &                       & Better & Equal & Worse & Gap  & Better & Equal & Worse & Gap   & Better & Equal & Worse & Gap  \\ 
\hline
R1                    & \multirow{6}{*}{2}    & 12     & 0     & 0     & -3.22 & 9      & 0     & 3     & -0.82 & 11     & 0     & 1     & -2.30  \\ 
\cline{1-1}\cline{3-14}
R2                    &                       & 11     & 0     & 0     & -2.49 & 1      & 0     & 10    & 10.18 & 1      & 0     & 10    & 9.69   \\ 
\cline{1-1}\cline{3-14}
C1                    &                       & 6      & 0     & 0     & -2.16 & 6      & 0     & 0     & -1.45 & 5      & 0     & 1     & -1.33  \\ 
\cline{1-1}\cline{3-14}
C2                    &                       & 2      & 6     & 0     & -0.83 & 1      & 3     & 4     & 2.72  & 1      & 3     & 4     & 2.72   \\ 
\cline{1-1}\cline{3-14}
RC1                   &                       & 7      & 1     & 0     & -0.79 & 3      & 0     & 5     & 5.24  & 5      & 0     & 3     & 1.05   \\ 
\cline{1-1}\cline{3-14}
RC2                   &                       & 4      & 3     & 1     & 0.12  & 0      & 0     & 8     & 9.77  & 0      & 0     & 8     & 10.92  \\ 
\hline
R1                    & \multirow{6}{*}{7}    & 12     & 0     & 0     & -4.59 & 7      & 0     & 5     & 6.36  & 11     & 0     & 1     & -2.37  \\ 
\cline{1-1}\cline{3-14}
R2                    &                       & 8      & 0     & 3     & -1.98 & 0      & 0     & 11    & 13.41 & 0      & 0     & 11    & 9.57  \\ 
\cline{1-1}\cline{3-14}
C1                    &                       & 6      & 3     & 0     & -2.25 & 5      & 0     & 4     & -1.68 & 5      & 0     & 4     & -1.95  \\ 
\cline{1-1}\cline{3-14}
C2                    &                       & 3      & 5     & 0     & -0.49 & 2      & 3     & 3     & 4.00  & 2      & 3     & 3     & 4.00   \\ 
\cline{1-1}\cline{3-14}
RC1                   &                       & 8      & 0     & 0     & -2.80 & 5      & 0     & 3     & 3.53  & 6      & 0     & 2     & -0.95  \\ 
\cline{1-1}\cline{3-14}
RC2                   &                       & 7      & 0     & 1     & -2.93 & 1      & 0     & 7     & 9.93  & 0      & 0     & 8     & 8.95   \\ 
\hline
R1                    & \multirow{6}{*}{13}   & 12     & 0     & 0     & -3.82 & 6      & 0     & 6     & 7.10  & 8      & 0     & 4     & -1.02  \\ 
\cline{1-1}\cline{3-14}
R2                    &                       & 9      & 1     & 1     & -2.74 & 0      & 0     & 11    & 12.14 & 0      & 0     & 11    & 8.60   \\ 
\cline{1-1}\cline{3-14}
C1                    &                       & 5      & 3     & 1     & -1.78 & 3      & 0     & 6     & 1.42  & 4      & 0     & 5     & -0.48  \\ 
\cline{1-1}\cline{3-14}
C2                    &                       & 2      & 6     & 0     & -1.84 & 4      & 0     & 4     & 3.23  & 4      & 0     & 4     & 2.84   \\ 
\cline{1-1}\cline{3-14}
RC1                   &                       & 8      & 0     & 0     & -1.72 & 3      & 0     & 5     & 4.58  & 4      & 0     & 4     & 3.35   \\ 
\cline{1-1}\cline{3-14}
RC2                   &                       & 6      & 2     & 0     & -2.25 & 0      & 0     & 8     & 15.03 & 0      & 0     & 8     & 14.60  \\
\hline
\end{tabular}
\caption{CP and lp-ALNS vs cp-ALNS on 25-customer instances}
\label{tab1}
\end{table}

The second experiment is to investigate the performance of the lp-ALNS on all the instances. The computational results are summarized in Figures \ref{fig:objective} and \ref{fig:bks}, and Table \ref{tab2} in Appendix. Figure \ref{fig:objective} shows that our algorithm clearly dominates the cp-ALNS in terms of solution quality. It provides better than average results on all instance classes. Figure \ref{fig:bks} reports the number of new best-known solutions found by the lp-ALNS for each class of instances. In total, 620 best-known solutions have been improved by our lp-ALNS.

Moreover, Table \ref{tab2} in the Appendix shows that the improvement on the value of the objective function created by the lp-ALNS is significant, especially on large instances (up to 16.75~\%). The relatively high gap between final solutions and initial solutions shows the efficiency of construction and deconstruction operators. However, similar to cp-ALNS, the computation time of our algorithm is still high. It depends heavily on the number of customers $|V|$ and that of special customers $|V_s|$. More specifically, in these cases, the number of variables and constraints in the LP models increase rapidly, leading to larger programs which are harder to solve.  

\begin{figure}[H]
  \centering
  \includegraphics[scale = 0.125]{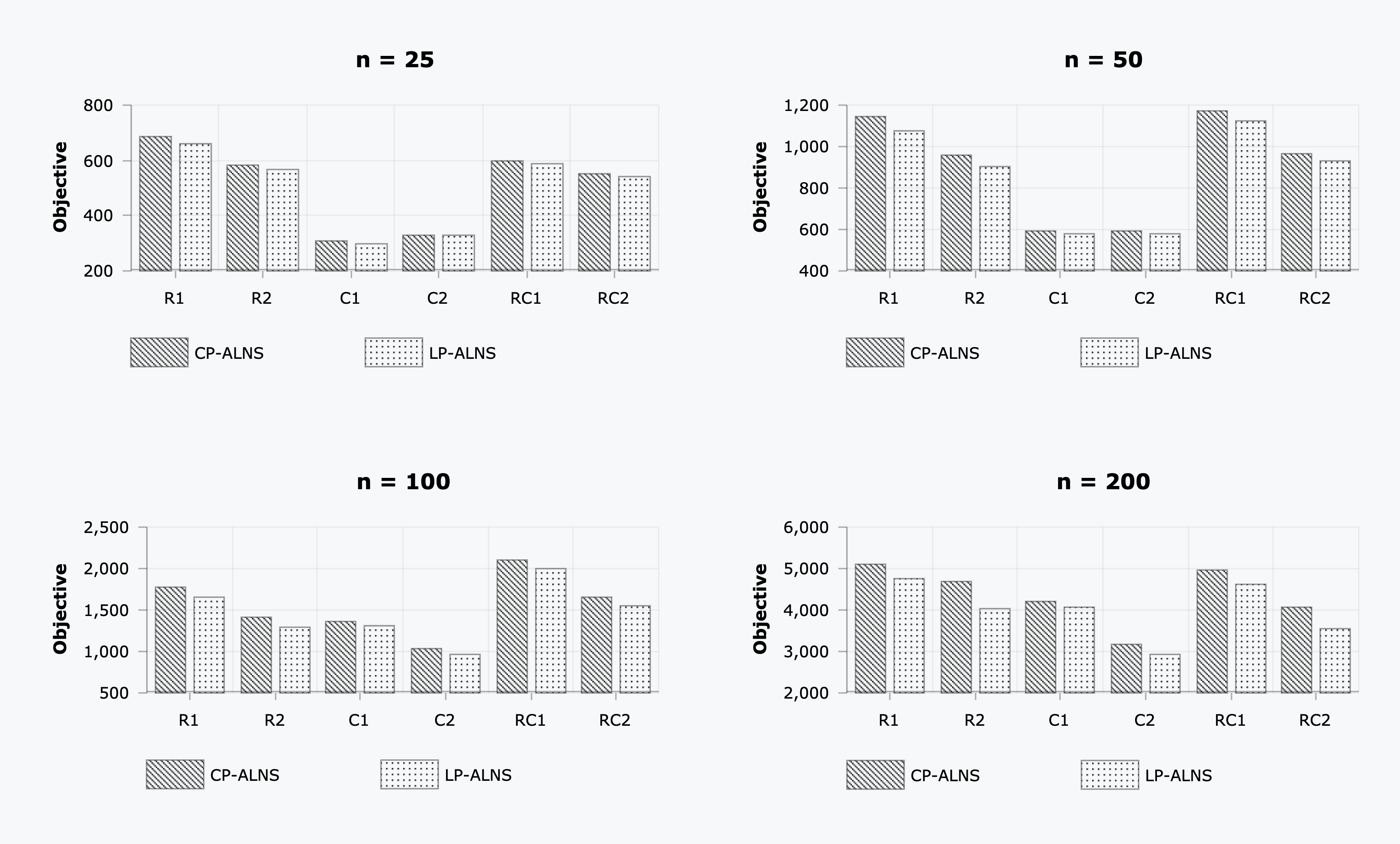}
  \caption{Comparison between lp-ALNS and cp-ALNS in terms of objective values on average}
  \label{fig:objective}
\end{figure}

\begin{figure}[H]
  \centering
  \includegraphics[scale = 0.25]{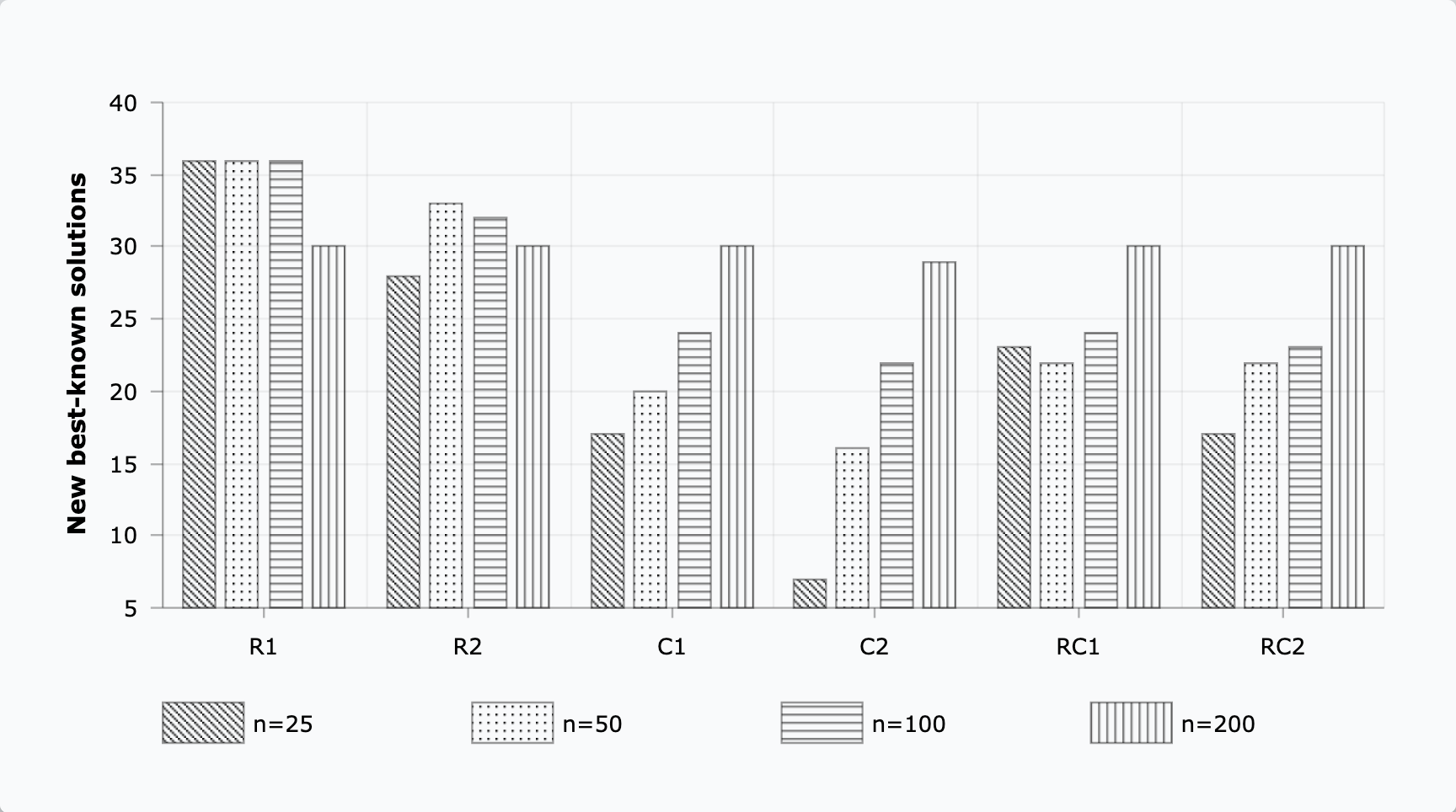}
  \caption{Number of new best-known solutions found by the lp-ALNS}
  \label{fig:bks}
\end{figure}

\section{Conclusion}
\label{conclude}
In this research, we study an important variant of the vehicle routing problem with synchronization constraints, which has numerous real-world applications. We propose a new CP model and an ALNS algorithm. The most remarkable feature of our ALNS is that we use linear programming to check the feasibility of insertions. A number of acceleration techniques have been proposed to significantly reduce the computation time of the algorithm. The obtained results on the benchmark instances from the literature show the performance of our method. Our CP model can even provide better solutions than the CP-based metaheuristic for small instances in much shorter running time. Our lp-ALNS dominates the cp-ALNS, in terms of solution quality, when it improves 620 best known solutions over 681 instances, and the improvement gap is relatively high.

The research perspectives are numerous. First, although our CP model provides very good solutions on small instances, it cannot prove any of them optimal. This, combined with the fact that there is no efficient exact method so far to solve the VRP with synchronization constraint, proves the hardness of this class of problem. An efficient and exact method is still an open question. Second, we believe that our lp-ALNS can be a used as a general framework, as it is easy to incorporate other constraints into the LP models. Thus, applying our method to solve other hard variants of VRPs could be an interesting research direction. Finally, our lp-ALNS is still quite time-consuming. Other acceleration techniques exploiting special structures of the LP programs which validate the insertion feasibility are required to make the algorithm become an efficient general solver for VRPs with rich attributes.

\appendix
\section*{Appendix}
Table \ref{tab2} reports the comparison between lp-ALNS and cp-ALNS in terms of objective values on average. The three first columns represent the size, class name, and the number of special customers of each instance class. Columns 4 and 5 report the objective of solutions (Column ``FinalObj") and running time in seconds (Column ``RunTime") on average of the cp-ALNS. Next columns show the results on average for each instance class of our lp-ALNS. More precisely, ``InitialObj" is the objective value of the initial solution constructed by regret-2 heuristic. ``FinalObj" is the objective value of the final solution after the search is stopped. ``RunTime" reports the computational time in seconds. Column ``Imp\%" shows the improvement of final solutions compared with initial ones. The final column ``gap\%" presents the gap between lp-ANLS and cp-ALNS solutions. A negative value means lp-ALNS provides a better solution.

\begin{table}[ht]
\centering
\setlength{\tabcolsep}{12pt}
\renewcommand{\arraystretch}{0.75}
\begin{adjustbox}{width=\textwidth, height=0.5\textheight}
\begin{tabular}{|ccc|cc|cccc|c|}\hline
%\multicolumn{1}{l}{} & \multicolumn{1}{l}{} & \multicolumn{1}{l}{} & \multicolumn{2}{l}{}               & \multicolumn{4}{l}{}                       & \multicolumn{1}{l}{}  \\
\multicolumn{3}{|c|}{Data} &                       \multicolumn{2}{c|}{cp-ALNS} & \multicolumn{4}{c|}{lp-ALNS}                    & \multirow{2}{*}{gap (\%)}  \\ 
\cline{1-9}
Size                 & Class                & Sync                  & FinalObj & RunTime                 & InitialObj & FinalObj & RunTime  & Imp\%   &                 \\ 
\hline
\multirow{6}{*}{25}  & R1                   & \multirow{6}{*}{2}   & 544.2487 & 8406.7                  & 667.7745   & 526.9739 & 1347.8   & 20.86\% & -3.22\%               \\
                     & R2                   &                      & 458.2109 & 9978.2                  & 594.1048   & 446.6154 & 1319.3   & 24.42\% & -2.49\%               \\
                     & C1                   &                      & 246.9088 & 9255.3                 & 371.2795    & 241.5390 & 1292.7   & 32.26\% & -2.16\%               \\
                     & C2                   &                      & 286.4098 & 7643.6                  & 342.2520   & 283.9314 & 1415.9   & 16.63\% & -0.83\%               \\
                     & RC1                  &                      & 520.1329 & 6805.4                  & 584.4385   & 515.7725 & 1480.5   & 11.23\% & -0.79\%               \\
                     & RC2                  &                      & 485.1764 & 7672.3                  & 624.6635   & 485.7430 & 1650.5   & 21.82\% & 0.12\%                \\ 
\hline
\multirow{6}{*}{25}  & R1                   & \multirow{6}{*}{7}   & 692.9658 & 7846.3                  & 888.4291   & 661.7204 & 2697.3   & 25.68\% & -4.59\%               \\
                     & R2                   &                      & 587.3948 & 10289.6                 & 753.2626   & 575.4996 & 3011.7   & 22.83\% & -1.98\%               \\
                     & C1                   &                      & 318.3507 & 8304.2                  & 485.0793   & 311.0783 & 2489.7   & 34.81\% & -2.25\%               \\
                     & C2                   &                      & 325.0070 & 7308.5                  & 411.1374   & 323.3936 & 2090.1   & 20.95\% & -0.49\%               \\
                     & RC1                  &                      & 623.5873 & 6388.2                  & 900.0518   & 605.3236 & 2707.9   & 31.69\% & -2.80\%               \\
                     & RC2                  &                      & 573.4323 & 8828.2                  & 796.9595   & 556.5850 & 2987.9   & 29.59\% & -2.93\%               \\ 
\hline
\multirow{6}{*}{25}  & R1                   & \multirow{6}{*}{13}  & 821.1631 & 7683.6                  & 1113.333   & 790.7901 & 3989.3   & 28.71\% & -3.82\%               \\
                     & R2                   &                      & 701.3905 & 10194.9                 & 930.8132   & 681.8161 & 4523.5   & 26.21\% & -2.74\%               \\
                     & C1                   &                      & 356.2599 & 6594.2                  & 585.0626   & 349.7486 & 3481.7   & 39.33\% & -1.78\%               \\
                     & C2                   &                      & 384.0774 & 6977.4                  & 461.5924   & 376.7076 & 3511.1   & 17.86\% & -1.84\%               \\
                     & RC1                  &                      & 657.5279 & 5892.3                  & 1009.331   & 644.6960 & 3844.1   & 35.93\% & -1.72\%               \\
                     & RC2                  &                      & 600.0163 & 8504.3                  & 905.7041   & 586.7275 & 4460.4   & 33.84\% & -2.25\%               \\ 
\hline
\multirow{6}{*}{50}  & R1                   & \multirow{6}{*}{3}   & 909.5803 & 19518.2                 & 1143.763   & 860.4969 & 4508.1   & 24.75\% & -5.69\%               \\
                     & R2                   &                      & 747.6849 & 23888.9                 & 999.7746   & 710.4967 & 4496.9   & 28.77\% & -5.03\%               \\
                     & C1                   &                      & 460.1213 & 39268.2                 & 554.8499   & 445.4759 & 4532.7   & 19.14\% & -3.13\%               \\
                     & C2                   &                      & 476.6964 & 35246.2                 & 633.3015   & 466.1748 & 4804.5   & 25.20\% & -2.02\%               \\
                     & RC1                  &                      & 959.1396 & 17810.0                 & 1165.249   & 932.1801 & 4539.8   & 19.44\% & -2.98\%               \\
                     & RC2                  &                      & 788.3569 & 20046.5                 & 1201.248   & 775.9619 & 4857.4   & 35.07\% & -1.65\%               \\ 
\hline
\multirow{6}{*}{50}  & R1                   & \multirow{6}{*}{13}  & 1178.034 & 19893.2                 & 1508.424   & 1103.480 & 10834.1  & 26.81\% & -6.57\%               \\
                     & R2                   &                      & 991.7101 & 23972.1                 & 1352.777   & 931.9242 & 13253.3  & 30.88\% & -6.03\%               \\
                     & C1                   &                      & 625.0501 & 18575.3                 & 951.2536   & 607.6679 & 9646.9   & 35.13\% & -2.70\%               \\
                     & C2                   &                      & 610.4964 & 22449.2                 & 907.3918   & 603.0134 & 13477.1  & 31.94\% & -1.20\%               \\
                     & RC1                  &                      & 1231.836 & 17850.4                 & 1664.564   & 1167.640 & 11213.3  & 29.66\% & -5.15\%               \\
                     & RC2                  &                      & 1014.296 & 22600.4                 & 1629.658   & 975.5894 & 12489.1  & 39.79\% & -3.96\%               \\ 
\hline
\multirow{6}{*}{50}  & R1                   & \multirow{6}{*}{25}  & 1354.380 & 19456.9                 & 1786.241   & 1259.770 & 18880.4  & 29.63\% & -7.15\%               \\
                     & R2                   &                      & 1134.506 & 24311.6                 & 1580.263   & 1057.765 & 26470.6  & 32.82\% & -6.96\%               \\
                     & C1                   &                      & 689.7339 & 16392.7                 & 1100.512   & 674.8498 & 18270.7  & 38.23\% & -2.09\%               \\
                     & C2                   &                      & 697.0899 & 20694.2                 & 1069.437   & 668.8730 & 23789.4  & 35.90\% & -3.85\%               \\
                     & RC1                  &                      & 1331.455 & 17536.3                 & 2064.870   & 1270.645 & 18259.1  & 38.31\% & -4.30\%               \\
                     & RC2                  &                      & 1087.594 & 23088.4                 & 1858.651   & 1041.364 & 21808.5  & 43.37\% & -4.50\%               \\ 
\hline
\multirow{6}{*}{100} & R1                   & \multirow{6}{*}{5}   & 1429.048 & 82716.3                 & 1847.463   & 1349.893 & 18128.8  & 26.94\% & -5.87\%               \\
                     & R2                   &                      & 1114.927 & 107910.0                & 1587.720   & 1038.766 & 19082.2  & 34.41\% & -6.98\%               \\
                     & C1                   &                      & 1017.524 & 41401.6                 & 1493.110   & 1005.144 & 21153.7  & 32.45\% & -1.20\%               \\
                     & C2                   &                      & 841.6708 & 75868.3                 & 1156.707   & 792.8305 & 20527.3  & 30.57\% & -5.74\%               \\
                     & RC1                  &                      & 1637.621 & 73411.3                 & 2187.973   & 1575.960 & 19985.8  & 27.97\% & -3.86\%               \\
                     & RC2                  &                      & 1286.366 & 97137.4                 & 1983.771   & 1219.061 & 20101.6  & 38.63\% & -5.47\%               \\ 
\hline
\multirow{6}{*}{100} & R1                   & \multirow{6}{*}{25}  & 1789.477 & 80930.0                 & 2353.730   & 1667.065 & 39167.0  & 29.28\% & -7.01\%               \\
                     & R2                   &                      & 1407.660 & 106302.7                & 2108.117   & 1303.016 & 55645.6  & 37.91\% & -7.66\%               \\
                     & C1                   &                      & 1441.884 & 74134.6                 & 2335.434   & 1396.531 & 38251.0  & 39.86\% & -3.13\%               \\
                     & C2                   &                      & 1087.373 & 70059.3                 & 1670.794   & 1004.834 & 39562.4  & 38.45\% & -7.24\%               \\
                     & RC1                  &                      & 2146.339 & 71074.1                 & 2985.869   & 2045.495 & 40795.9  & 31.44\% & -4.73\%               \\
                     & RC2                  &                      & 1709.736 & 99529.7                 & 2594.230   & 1610.915 & 51314.6  & 37.95\% & -6.06\%               \\ 
\hline
\multirow{6}{*}{100} & R1                   & \multirow{6}{*}{50}  & 2092.569 & 72804.5                 & 2831.534   & 1931.541 & 59256.5  & 32.07\% & -7.91\%               \\
                     & R2                   &                      & 1693.735 & 104309.4                & 2526.176   & 1513.136 & 89195.5  & 40.16\% & -10.91\%              \\
                     & C1                   &                      & 1609.269 & 72829.5                 & 2911.811   & 1545.038 & 57106.8  & 46.57\% & -3.91\%               \\
                     & C2                   &                      & 1180.336 & 54500.8                 & 1927.523   & 1078.205 & 69467.0  & 41.82\% & -8.42\%               \\
                     & RC1                  &                      & 2522.189 & 67991.7                 & 3650.564   & 2361.475 & 60008.8  & 35.39\% & -6.45\%               \\
                     & RC2                  &                      & 1945.313 & 95459.4                 & 3173.071   & 1817.931 & 63274.9  & 42.44\% & -6.83\%               \\ 
\hline
\multirow{6}{*}{200} & R1                   & \multirow{6}{*}{10}  & 4144.381 & 126805.8                & 5674.585   & 3893.875 & 37843.3  & 31.74\% & -6.55\%               \\
                     & R2                   &                      & 3713.917 & 161933.9                & 5317.305   & 3320.119 & 37547.4  & 37.72\% & -11.05\%              \\
                     & C1                   &                      & 3391.796 & 100174.6                & 5108.380   & 3299.623 & 39677.3  & 34.62\% & -2.67\%               \\
                     & C2                   &                      & 2578.896 & 145855.5                & 3861.907   & 2364.314 & 36008.3  & 38.13\% & -8.26\%               \\
                     & RC1                  &                      & 4088.110 & 138911.3                & 5642.754   & 3884.939 & 38358.8  & 31.03\% & -4.95\%               \\
                     & RC2                  &                      & 3303.763 & 149639.7                & 4898.392   & 2917.336 & 39697.6  & 40.55\% & -11.91\%              \\ 
\hline
\multirow{6}{*}{200} & R1                   & \multirow{6}{*}{50}  & 5074.646 & 136440.0                & 7240.237   & 4724.183 & 74670.6  & 35.08\% & -7.46\%               \\
                     & R2                   &                      & 4769.591 & 163572.0                & 6608.872   & 4074.128 & 93781.0  & 38.47\% & -14.99\%              \\
                     & C1                   &                      & 4238.077 & 120709.5                & 7290.512   & 4126.756 & 74785.5  & 43.11\% & -2.65\%               \\
                     & C2                   &                      & 3254.876 & 161222.8                & 5851.064   & 3008.252 & 81829.4  & 48.25\% & -7.49\%               \\
                     & RC1                  &                      & 4944.824 & 144934.9                & 7017.894   & 4573.909 & 78424.0  & 34.65\% & -7.50\%               \\
                     & RC2                  &                      & 4039.977 & 155597.6                & 6108.987   & 3539.513 & 92402.6  & 41.75\% & -12.49\%              \\ 
\hline
\multirow{6}{*}{200} & R1                   & \multirow{6}{*}{100} & 6139.267 & 137336.0                & 8718.394   & 5704.425 & 116804.2 & 34.72\% & -7.55\%               \\
                     & R2                   &                      & 5616.638 & 157111.4                & 7993.287   & 4699.330 & 154303.7 & 41.40\% & -16.75\%              \\
                     & C1                   &                      & 4964.759 & 111991.6                & 9117.560   & 4757.915 & 112905.5 & 47.57\% & -4.16\%               \\
                     & C2                   &                      & 3646.121 & 132923.9                & 6969.890   & 3374.519 & 146172.7 & 50.71\% & -7.31\%               \\
                     & RC1                  &                      & 5836.127 & 148733.9                & 8265.386   & 5395.134 & 122680.9 & 34.72\% & -7.58\%               \\
                     & RC2                  &                      & 4882.749 & 159016.6                & 7138.363   & 4204.200 & 153580.6 & 41.12\% & -13.99\%              \\
\hline
\end{tabular}
\end{adjustbox}
\caption{Comparison between lp-ALNS and cp-ALNS}
\label{tab2}
\end{table}


\begin{thebibliography}{99}
%\bibitem{CTSP}
%J-A. Chisman. The clustered traveling salesman problem. {\em Computers \& Operations Research} 2:2 (1975), 115-119.
\bibitem{drexl}
Drexl, M. (2012). \textit{Synchronization in vehicle routing—a survey of VRPs with multiple synchronization constraints}. Transportation Science 46.3: 297-316.

\bibitem{rasmussen}
Rasmussen, M.-S., Justesen, T., Dohn, A., \& Larsen, J. (2012). \textit{The home care crew scheduling problem: Preference-based visit clustering and temporal dependencies}. European Journal of Operational Research 219.3: 598-610.

\bibitem{rousseau}
Rousseau, L.-M., Gendreau, M., \& Pesant, G. (2013). \textit{The synchronized dynamic vehicle dispatching problem}. INFOR: Information Systems and Operational Research 51.2: 76-83.

\bibitem{bredstrom}
Bredström, D., \& Rönnqvist, M. (2008). \textit{Combined vehicle routing and scheduling with temporal precedence and synchronization constraints.} European journal of operational research 191.1: 19-31.

\bibitem{afifi}
Afifi, S., Dang, D-C., \& Moukrim, A. (2016). \textit{Heuristic solutions for the vehicle routing problem with time windows and synchronized visits}. Optimization Letters 10.3: 511-525.

\bibitem{pillac}
Pillac, V., Gu\'eret, C., \& Medaglia, A. L. (2018). \textit{A fast reoptimization approach for the dynamic technician routing and scheduling problem.} Recent Developments in Metaheuristics. Springer, Cham, 347-367.

\bibitem{parragh}
Parragh, S-N., \& Doerner, K-F. (2018). \textit{Solving routing problems with pairwise synchronization constraints}. Central European journal of operations research 26.2: 443-464.

\bibitem{Hojabri}
Hojabri, H., Gendreau, M., Potvin, J-Y., \& Rousseau, L.-M. (2018). \textit{Large neighborhood search with constraint programming for a vehicle routing problem with synchronization constraints}. Computers \& Operations Research 92, 87-97.

\bibitem{CPO} 
IBM Software, (2015). \textit{IBM ILOG CPLEX Optimization Studio V12.6.3}.

\bibitem{Laborie2009}
Laborie, P. (2009). \textit{IBM ILOG CP Optimizer for detailed scheduling illustrated on three problems}. In International Conference on AI and OR Techniques in Constraint Programming for Combinatorial Optimization Problems (pp. 148-162). Springer, Berlin, Heidelberg.

\bibitem{Ghedira2013}
Gh\'edira, K. (2013). \textit{Constraint satisfaction problems: CSP formalisms and techniques}. John Wiley \& Sons.
\bibitem{Goel2015}
Goel, V., Slusky, M., van Hoeve, W.J., Furman, K.C., \& Shao, Y. (2015). \textit{Constraint programming for LNG ship scheduling and inventory management}. European Journal of Operational Research, 241(3), pp. 662-673.

\bibitem{Ham2018} 
Ham, A.M. (2018). \textit{Integrated scheduling of m-truck, m-drone, and m-depot constrained by time-window, drop-pickup, and m-visit using constraint programming}. Transportation Research Part C: Emerging Technologies 91, 1-14.

\bibitem{Ropke}
Pisinger, D. \& R\o pke, S. (2007). \textit{A general heuristic for vehicle routing problems}. Computers \& Operations Research 34: 2403-2435. 

\bibitem{Kindervater}
Kindervater, G.A.P.. \&  Savelsbergh M.W.P.  (1997). \textit{Vehicle routing: handling edge exchanges}. in Aarts E.H.L., Lenstra J.K. (eds.), Local Search in Combinatorial Optimization, John Wiley \& Sons, pp. 337–360.

\bibitem{Solomon1987}
Solomon, M.M. (1987). \textit{Algorithms for the vehicle routing and scheduling problems with time window constraints}. Operations Research 35, 254–265.

\bibitem{Homberger1999}
Homberger, J., \& Gehring, H. (1999). \textit{Two evolutionary metaheuristics for the vehicle routing problem with time windows}. INFOR 37, 297–318.

\bibitem{Pesant}
Pesant, G., Gendreau, M., Potvin, J.-Y., Rousseau, J.-M. (1998). \textit{An Exact Constraint Logic Programming Algorithm for the Traveling Salesman Problem with Time Windows}. Transportation
Science, 32: 12–29.

\bibitem{Shaw}
Shaw P. (1998). \textit{Using Constraint Programming and Local Search Methods to Solve Vehicle Routing Problems}. In Principles and Practice of Constraint Programming, Goos, G., Hartmanis, J., van Leeuwen, J., eds., Lecture Notes in Computer Science 1520, Springer, Berlin, pp. 417-431.


\end{thebibliography}
\end{document}